\documentclass{article}

\usepackage{microtype}
\usepackage{graphicx}
\usepackage{subcaption}
\usepackage{booktabs}
\usepackage{multirow}
\usepackage{hyperref}

\usepackage[preprint]{icml2026}

\usepackage{amsmath}
\usepackage{amssymb}
\usepackage{mathtools}
\usepackage{amsthm}

\usepackage[capitalize,noabbrev]{cleveref}

\theoremstyle{plain}

\theoremstyle{definition}

\theoremstyle{remark}

\icmltitlerunning{FSA}

\begin{document}

\twocolumn[
  \icmltitle{Feature to Dynamics: Feature-space to Autoregression strategy for Zero-shot Time Series Forecasting }

  \icmlsetsymbol{equal}{*}

  \begin{icmlauthorlist}
    \icmlauthor{Yifan Wu}{ai}
    \icmlauthor{Junjie Wu}{ai}
    \icmlauthor{Kai Wu\textsuperscript{*}}{ai}
    \icmlauthor{Xiaoyu Zhang}{cyber}
    \icmlauthor{Jian Lou}{sysu}
  \end{icmlauthorlist}

  \icmlaffiliation{ai}{School of Artificial Intelligence, Xidian University, Xi'an, China}
  \icmlaffiliation{cyber}{School of Cyber Engineering, Xidian University, Xi'an, China}
  \icmlaffiliation{sysu}{Sun Yat-Sen University, Guangzhou, China}

  \icmlcorrespondingauthor{Kai Wu}{kwu@xidian.edu.cn}

  \icmlkeywords{Machine Learning, ICML}

  \vskip 0.3in
]

\printAffiliationsAndNotice{}

\begin{abstract}
Zero-shot time series forecasting aims to predict future values for previously unseen series, requiring models to generalize temporal dynamics beyond the training distribution. While recent foundation models achieve strong in-domain performance through large-scale pretraining, their effectiveness often relies on broad data coverage and implicit pattern memorization, which can limit generalization when data are scarce or source and target domains are disjoint. In this work, we propose FSA, a feature-to-strategy framework for controlled zero-shot univariate forecasting. Instead of directly modeling raw sequences in the observation space, FSA learns a structured mapping from an interpretable feature space to an autoregressive strategy space. This design introduces explicit inductive biases that disentangle global trends, periodic components, and local temporal dynamics, enabling the model to capture transferable time-series structure with fewer data assumptions. Empirical results show that, under identical pretraining data, training protocol, and comparable parameter budgets, FSA outperforms Transformer-based architectures in our controlled zero-shot setting.

\end{abstract}

\section{Introduction}

Time series forecasting is a fundamental pillar of modern decision-making, enabling resilience in systems ranging from energy grids to global logistics~\cite{hu2022times,TSOLAKI2023284}. Recently, the pursuit of zero-shot forecasting—the ability to generalize to unseen dynamics without task-specific fine-tuning—has become the primary frontier. This ambition has birthed Time Series Foundation Models (TSFMs), which seek to internalize universal temporal patterns from massive, heterogeneous datasets.
\begin{figure}
    \centering
    \begin{subfigure}{0.22\textwidth}
    \includegraphics[width=0.9\linewidth]{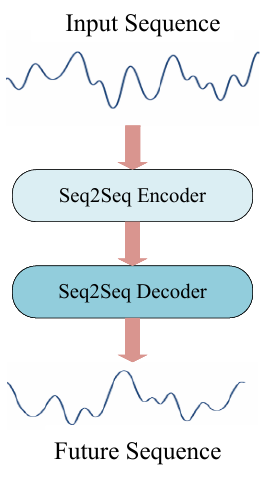}
    \caption{Seq2Seq}
    \label{Seq2seq}
    \end{subfigure}
    \hfill
    \begin{subfigure}{0.22\textwidth}
    \includegraphics[width=0.9\linewidth]{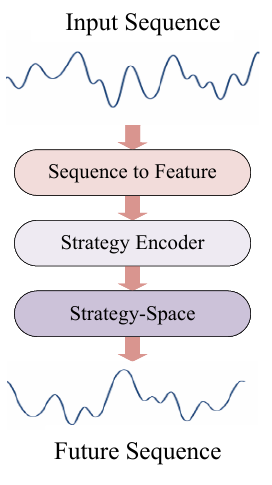}
    \caption{Feature to Strategy}
    \label{Feature to Strategy}
    \end{subfigure}
    \caption{Paradigm shift: From the Seq2Seq pattern of time-series models to our proposed mapping from feature space to strategy space. (a) The traditional sequence-to-sequence (Seq2Seq) paradigm for time-series forecasting; (b) Our proposed feature space to strategy space paradigm. }
    \vskip -0.1in
    \label{fig:placeholder}
\end{figure}

However, the prevailing architecture of many TSFMs remains tethered to the sequence-to-sequence (Seq2Seq) paradigm. As shown in Figure \ref{Seq2seq}, this approach treats forecasting as a high-dimensional mapping problem, where neural networks are trained to translate an input history sequence ($X$) into a precise future sequence ($Y$). While computationally powerful, this paradigm can require broad coverage of source domains to transfer reliably. In a controlled zero-shot setting where pretraining uses only a limited set of source domains, it encounters two practical issues:
\begin{itemize}
    \item \textbf{Limited Source Coverage}: The space of possible time series is extremely broad. A compact pretraining corpus cannot cover all potential combinations of trend, seasonality, noise, and local dynamics.
    \item \textbf{Value-Space Overfitting}: Directly fitting future values can encourage models to memorize dataset-specific surface patterns when training diversity is limited, rather than emphasizing reusable forecasting rules.
\end{itemize}
In this paper, we propose FSA (Feature-to-Strategy Adaptation), a novel framework that navigates these obstacles by shifting the objective from predicting values to generating strategies(See Figure \ref{Feature to Strategy}). Our approach is inspired by human cognitive flexibility: when faced with a novel problem, humans do not leap directly to an optimal numerical solution. Instead, we distill the problem's characteristics and iteratively refine a strategy to solve it. To address the aforementioned challenges, FSA introduces two strategic design shifts:
\begin{itemize}
    \item \textbf{From Values to Strategies (Overcoming Target Scarcity)}: Instead of forcing the model to output exact future points (the ``what''), FSA generates the parameters of an autoregressive strategy (the ``how''). By predicting the rules that govern the future rather than the future itself, we circumvent the difficulty of collecting perfect target sequences and impose a stronger inductive bias for temporal consistency.
    \item \textbf{From Sequences to Task Features (Reducing Source Coverage Dependence)}: To avoid relying on raw-sequence coverage alone, we construct a condensed Task Feature Space. This space distills high-level methodological information--such as structural invariants and local-global dependencies--that generalizes across tasks. This allows the model to recognize the structure of a new time series even if it has never seen that specific sequence before.
\end{itemize}

Following this philosophy, our implementation utilizes a dual-stage architecture: a feature extractor that maps raw observations into a task-aware latent space, and a strategy generator that outputs optimal autoregressive coefficients.

It is important to clarify that our goal is not to present a larger industrial-scale foundation model, but to test whether a feature-to-strategy architecture provides stronger zero-shot transfer under controlled data and capacity constraints. By keeping training data and computational budgets identical, we compare FSA against capacity-matched sequence models on disjoint target benchmarks. This study therefore supports an architectural claim: structured strategy generation can be a data-efficient alternative to direct value prediction in the low-to-moderate capacity regime.

Our primary contributions are as follows:
\begin{itemize}
    \item We introduce the Feature-to-Strategy learning objective, challenging the long-standing Seq2Seq dominance in zero-shot time series forecasting.
    \item We design the FSA framework, which utilizes a compact task feature space to generate autoregressive parameters, effectively decoupling task understanding from value generation.
    \item Through controlled experiments, we demonstrate that FSA exhibits stronger zero-shot generalization than standard Seq2Seq-based architectures under identical data and capacity constraints.
\end{itemize}

\section{Related work}

\paragraph{Statistical and Autoregressive Models}
Classical statistical models, most notably Autoregressive Integrated Moving Average (ARIMA)~\cite{box2015time}, have long been regarded as powerful tools for modeling such temporal structures. Automatic ARIMA fitting methods, such as AutoARIMA~\cite{hyndman2008automatic}, enable parameter estimation without manual intervention and achieve strong performance in short-horizon forecasting. However, these approaches implicitly assume that future dynamics are governed by the same distribution as the observed history, and they typically focus on one-step-ahead prediction. As a result, they struggle to generalize to multi-step forecasting scenarios or settings where the future dynamics deviate from historical patterns.

To overcome these limitations, a large body of research has explored learning-based approaches that extract latent representations from time series. Deep state space models~\cite{rangapuram2018deep} and autoregressive neural models such as DeepAR~\cite{salinas2020deepar} attempt to capture temporal dependencies through learned representations, improving flexibility over purely statistical methods. 

\paragraph{Time Series Foundation Models}

Recent developments in TSFMs have addressed these challenges by leveraging large-scale pretraining techniques. Early TSFMs largely adopt Transformer-based sequence-to-sequence or decoder-only architectures. For instance, TimesFM~\cite{das2024decoder} employs a decoder-only autoregressive Transformer with patch-based tokenization and demonstrates strong zero-shot performance on held-out datasets. Moirai~\cite{woo2024unified} proposes a masked-encoder framework with multi-scale patching and any-variate attention, enabling flexible handling of heterogeneous time series in zero-shot settings. Moirai 2.0~\cite{liu2025moirai} further enhances this approach with more efficient modeling and better generalization to diverse time series domains. More recently, scaling-oriented approaches such as TIME-MoE~\cite{shi2024time} leverage sparse mixture-of-experts architectures to increase model capacity while controlling inference cost, whereas Sundial~\cite{liu2025sundial} explores generative probabilistic objectives to improve uncertainty modeling under large-scale pretraining. Chronos 2.0~\cite{ansari2025chronos} extends the original Chronos framework by introducing advanced multi-scale temporal attention mechanisms to improve zero-shot capabilities across a range of forecasting tasks. TOTO~\cite{cohen2025time} develops a time series forecasting foundation model specialized for observability data, leveraging innovations like patch-based normalization and proportional time-variate factorized attention, which make it particularly effective for handling real-time system metrics and other complex time series data.

\paragraph{Tabular-Prior and Synthetic-Pretraining Forecasters}
Recent work has also explored forecasting through in-context priors and synthetic pretraining, including TempoPFN~\cite{moroshan2025tempopfn}, TabPFN-TS~\cite{hoo2025tables}, and TiRex~\cite{auer2026tirex}. These methods are closely related to our motivation because they also question whether zero-shot forecasting must rely only on direct sequence-to-sequence scaling. FSA differs in that it does not perform generic in-context posterior inference or generate the weights of a downstream neural model. Instead, it maps structured time-series descriptors to a compact, interpretable autoregressive forecasting strategy. We include additional comparisons to recent zero-shot baselines where compatible evaluation settings are available, and discuss the remaining gaps in Section~\ref{sec:limitations}.

\section{Preliminaries}

\textbf{Problem Definition}. In this work, we investigate the task of forecasting future values of a univariate time series under the zero-shot setting. Specifically, given a historical observation sequence of length \(T\), denoted as $\mathbf{X}_{1:T} = (x_1, x_2, \dots, x_T) \in \mathbb{R}^T$, 
the objective is to predict the subsequent \(H\) future values, expressed as $\hat{\mathbf{X}}_{T+1:T+H} = f_{\theta}(\mathbf{X}_{1:T}) \in \mathbb{R}^H$, where \(f_{\theta}\) represents a parameterized forecasting model and \(\hat{\mathbf{X}}_{T+1:T+H}\) denotes the predicted values over the forecast horizon \(H\).

\textbf{ARIMA}. ARIMA model is a fundamental class of linear stochastic processes used for modeling non-stationary time series. An $\text{ARIMA}(p, d, q)$ model is characterized by three parameters: the autoregressive order $p$, the degree of differencing $d$, and the moving average order $q$.

To handle non-stationarity, the differencing operator $\nabla$ is applied $d$ times to the original series $\mathbf{x}_{1:T}$ to obtain a stationary series $y_t = \nabla^d x_t$, where $\nabla x_t = x_t - x_{t-1}$. The resulting series $\{y_t\}$ is governed by the following linear equation:
\begin{equation}
    y_t = c + \sum_{i=1}^{p} \phi_i y_{t-i} + \sum_{j=1}^{q} \theta_j \varepsilon_{t-j} + \varepsilon_t,
\end{equation}
where $c$ is a constant intercept, $\{\phi_i\}_{i=1}^p$ are the AR coefficients, and $\{\theta_j\}_{j=1}^q$ are the MA coefficients. The term $\varepsilon_t$ denotes a white noise process, $\varepsilon_t \sim \text{WN}(0, \sigma^2)$, representing Gaussian innovations.

By integrating the AR component (capturing temporal dependencies) and the MA component (modeling innovation shocks), ARIMA provides a robust baseline for linear trend and seasonality modeling. Under the assumption of stationarity and invertibility, the model parameters can be estimated via maximum likelihood or least squares.

\section{Method}

\begin{figure*}
    \centering
    \includegraphics[
    width=0.9\linewidth,
    trim=3pt 3pt 3pt 3pt,
    clip
]{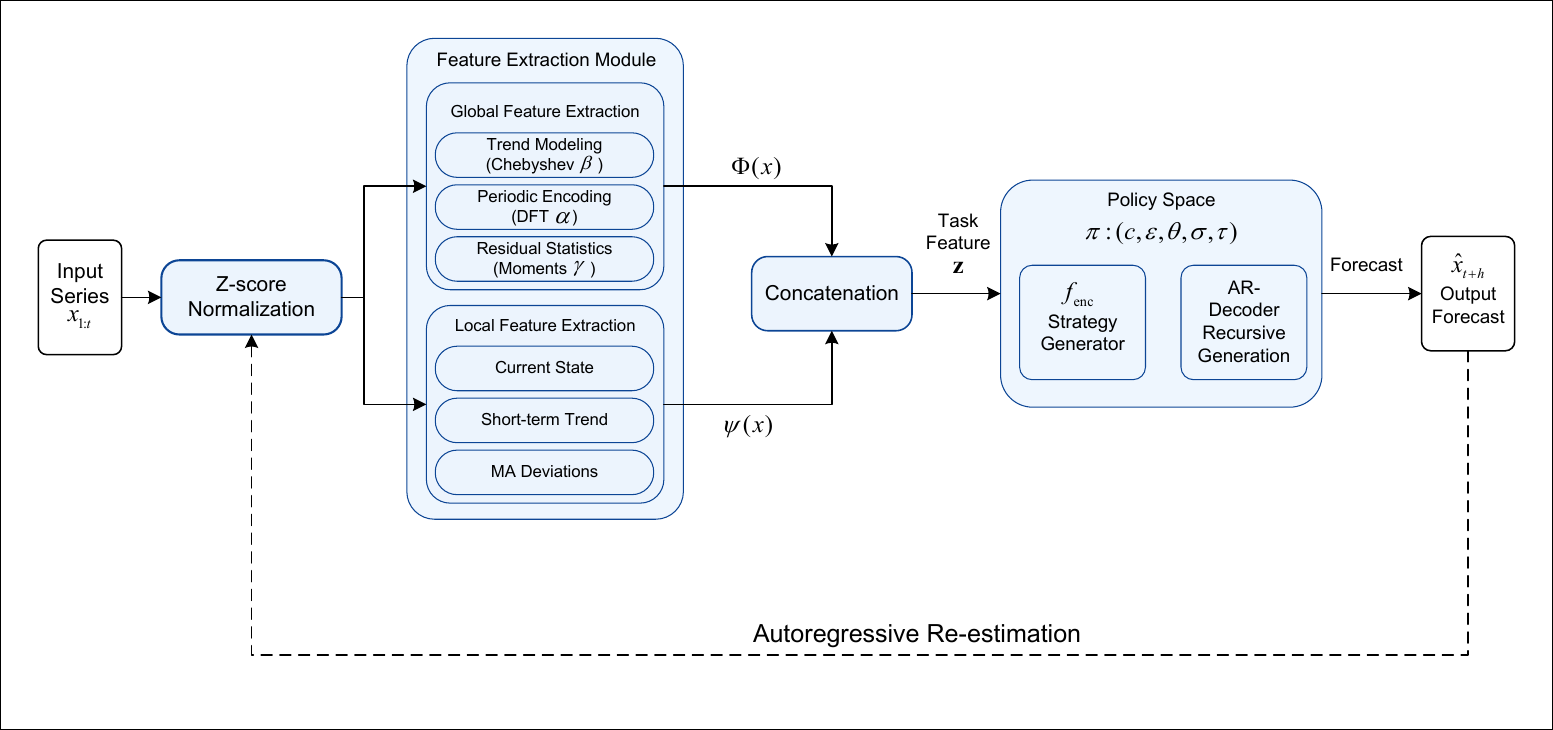}
    \caption{The overall framework of the proposed method.
    The architecture consists of two main stages. First, the Feature Extraction Module normalizes the input series $\mathbf{x}_{1:T}$ into $\tilde{\mathbf{x}}$, and extracts global structural features $\boldsymbol{\Phi}(\mathbf{x})$ (trend $\boldsymbol{\beta}$, periodicity $\boldsymbol{\alpha}$, residual statistics $\boldsymbol{\gamma}$) and local dynamic features $\boldsymbol{\psi}(\mathbf{x})$, concatenating them into a task feature vector $\mathbf{z}$. Second, in the Strategy Space, the Strategy Generator ($f_{\text{Enc}}$) maps $\mathbf{z}$ to AR-style strategy parameters $\boldsymbol{\pi} = (c, \boldsymbol{\phi}, \theta, \sigma)$. The recursive decoder then uses strategy $\boldsymbol{\pi}$ to generate future forecasts $\hat{x}_{T+h}$. The dashed line indicates the autoregressive re-estimation loop during inference, where generated predictions are fed back as pseudo-observations to update features for subsequent forecast segments.}
    \label{fig:fsa}
\end{figure*}

As shown in Figure~\ref{fig:fsa}, our implementation consists of a dual-stage architecture: a feature extraction module that captures structural invariants, and an MLP-based strategy generator that predicts parameters for an AR-style recursive decoder with an ARIMA-inspired parameterization. When a re-estimation step is triggered, the model autoregressively recomputes the sequence features based on the newly generated predictions and subsequently produces future forecasts.

\subsection{Design of Feature Space}

To reduce dependence on exhaustive source-domain coverage, we represent each time series through a combination of global structural descriptors and local dynamic signals.

\subsubsection{Global Feature Extraction}
Given an input series $\mathbf{x}_{1:T} = (x_1, \dots, x_T) \in \mathbb{R}^T$, we first apply z-score normalization to ensure scale-invariance. The normalized sequence $\tilde{\mathbf{x}}$ is decomposed into three components: $\tilde{\mathbf{x}} = \mathbf{g}(\boldsymbol{\beta}) + \mathbf{f}(\boldsymbol{\alpha}) + \mathbf{r}_{\text{final}}$, where $\mathbf{g}(\boldsymbol{\beta})$ captures the trend, $\mathbf{f}(\boldsymbol{\alpha})$ represents periodicity, and $\mathbf{r}_{\text{final}}$ denotes the stochastic residual.

\textbf{Trend Modeling:} We project the sequence onto a $B$-th order Chebyshev polynomial basis $\{T_n(t)\}_{n=0}^{B-1}$ defined on the normalized time interval $t \in [-1,1]$. The trend coefficients $\boldsymbol{\beta} = [\beta_0, \dots, \beta_{B-1}]$ are obtained via least-squares fitting, providing a numerically stable representation of long-term evolution.

\textbf{Periodic Encoding:} After detrending, we apply the Discrete Fourier Transform (DFT) to the residual $\mathbf{r}_{\text{trend}} = \tilde{\mathbf{x}} - \mathbf{g}(\boldsymbol{\beta})$. We select the top $A$ frequency bins with the highest magnitudes (excluding the DC component) to form the periodic feature $\boldsymbol{\alpha} = \{\hat{r}_{k_1}, \dots, \hat{r}_{k_A}\} \subset \mathbb{C}$. This captures the dominant cyclic behavior without modeling high-frequency noise.

\textbf{Residual Statistics:} The remaining component $\mathbf{r}_{\text{final}}$ is summarized by its low-order statistical moments:
\begin{equation}
\boldsymbol{\gamma} = \big[ \mathrm{Var}(\mathbf{r}_{\text{final}}), \mathrm{Skew}(\mathbf{r}_{\text{final}}), \mathrm{Kurt}(\mathbf{r}_{\text{final}}) \big], 
\end{equation}
which characterize the noise intensity, asymmetry, and tail behavior. 

The final global feature vector is defined as $\boldsymbol{\Phi}(\mathbf{x}) = [\boldsymbol{\beta}, \mathrm{Re}(\boldsymbol{\alpha}), \mathrm{Im}(\boldsymbol{\alpha}), \boldsymbol{\gamma}] \in \mathbb{R}^{B + 2A + 3}$, where $\mathrm{Re}(\boldsymbol{\alpha})$ and $\mathrm{Im}(\boldsymbol{\alpha})$ represent real and imaginary parts of $\boldsymbol{\alpha}$.

\subsubsection{Local Feature Extraction}
While global features describe structural invariants, short-term forecasting requires sensitivity to the most recent dynamics. We extract the following local descriptors from the terminal window of $\tilde{\mathbf{x}}$:
\begin{itemize}
    \item \textbf{Current State:} The last observed value $\tilde{x}_T$ and its relative position (min-max scaled within the window) to identify current levels relative to recent extremes.
    \item \textbf{Moving Average (MA) Deviations:} Inspired by technical analysis, we compute $\Delta_{k} = \tilde{x}_T - \text{MA}_k$ for $k \in \{10, 20\}$ to capture mean-reversion or momentum signals, where $\text{MA}_{k}$ refers to the moving average over the past $k$ points.
    \item \textbf{Short-term Trend:} The linear slope and endpoint difference \(\tilde{x}_t - \tilde{x}_{t-L+1}\) over the last $L$ points to indicate immediate directional bias.
\end{itemize}
These are concatenated into a local feature vector $\boldsymbol{\psi}(\mathbf{x})$. The total task feature is $\mathbf{z} = [\boldsymbol{\Phi}(\mathbf{x}), \boldsymbol{\psi}(\mathbf{x})]$.

\subsection{Design of Strategy Space}

\subsubsection{Strategy Generator (ARIMA-Encoder)}
Instead of predicting values, the encoder $f_{\text{Enc}}(\cdot)$ maps the task feature $\mathbf{z}$ to a strategy $\boldsymbol{\pi}$, which resides in the parameter space of an ARIMA$(p,d,q)$ model. 
In our implementation, we use a lightweight $(3,0,1)$ configuration. The encoder outputs:
\begin{equation}
(c, \boldsymbol{\phi}, \theta, \sigma) = f_{\text{Enc}}(\mathbf{z}),
\end{equation}
where $c$ is the drift, $\boldsymbol{\phi} \in \mathbb{R}^p$ are the AR coefficients, $\theta$ is the MA coefficient, and $\sigma$ is the innovation scale. The task feature dimension is 22, consisting of 16 global basis features and 6 local features. The strategy generator is a 3-layer MLP with hidden size 512, GELU activations, dropout of 0.1, and LayerNorm after each linear layer. Its output head produces
\begin{equation}
z_{\pi} \in \mathbb{R}^{p+q+2}, \quad p=3,\ q=1,
\end{equation}
which is parameterized as
\begin{equation}
\begin{aligned}
c &= z_{\pi,0}, \\
\phi_i &= \phi_{\max}\tanh(z_{\pi,i}), \\
\theta_1 &= \theta_{\max}\tanh(z_{\pi,4}), \\
\sigma &= z_{\pi,5}.
\end{aligned}
\end{equation}
with $\phi_{\max}=\theta_{\max}=0.96$. The output-layer weights are initialized with Xavier-uniform initialization using gain 0.01, and biases are initialized to $[0,0.1,0.1,0.1,0.05,0.01]$. The bounded $\tanh$ parameterization is a practical stabilization mechanism: it limits coefficient magnitudes and reduces the risk of explosive recursive rollout, but it is not a complete theoretical stationarity guarantee for general AR($p$) processes.

\subsubsection{AR-style Recursive Decoder}
The decoder treats the predicted parameters as a strategy for recursive generation. For $t > T$, the forecast $\hat{x}_t$ is generated as:
\begin{equation}
\hat{x}_t = c + \sum_{i=1}^{p} \phi_i \tilde{x}_{t-i}
\label{eq:ar_rollout}
\end{equation}
where $\tilde{x}_{t-i}$ is replaced by $\hat{x}_{t-i}$ as the horizon increases. Although the strategy head predicts an MA coefficient and innovation scale for an ARIMA-inspired parameterization, the deterministic rollout in Eq.~\ref{eq:ar_rollout} uses only the AR recursion and sets future innovations to their conditional mean. The decoder is therefore best described as an AR-style recursive decoder rather than a full stochastic ARIMA decoder. By outputting the "rules" of the series rather than the values, the model maintains temporal consistency even in zero-shot scenarios.

\subsection{Loss Function}
The FSA framework is trained to optimize the strategy generator by minimizing the discrepancy between the recursively generated trajectory and the ground truth. Although the model outputs an autoregressive strategy $\boldsymbol{\pi} = \{c, \boldsymbol{\phi}, \theta, \sigma\}$, we adopt a terminal-value-based Mean Squared Error (MSE) loss rather than a likelihood-based objective to ensure robustness against noise. 

Given the differentiable nature of the AR-style recursive decoder, we "unroll" the recursive generation process for a forecast horizon $H$. The loss function is defined as:
\begin{equation}
\mathcal{L}_{\text{MSE}} = \frac{1}{H} \sum_{h=1}^{H} \left\| \hat{x}_{t+h} - x_{t+h} \right\|^2
\end{equation}
where $\hat{x}_{t+h}$ is the value generated by the predicted strategy $\boldsymbol{\pi}$ at step $h$. During training, gradients are backpropagated through the recursive decoder into the strategy generator and feature extractor. This end-to-end flow forces the model to internalize structural invariants that lead to stable, long-term predictive policies rather than mere point-to-point mappings.

\subsection{Inference of FSA}
\subsubsection{Autoregressive Re-estimation}

To adapt to non-stationary dynamics, we employ a periodic re-estimation strategy. Rather than using a fixed strategy for the entire horizon $H$, the model generates a segment of length $K$. These predictions are then treated as pseudo-observations to update the local features $\boldsymbol{\psi}$ and global features $\boldsymbol{\Phi}$. 

A new forward pass through $f_{\text{Enc}}$ generates an updated strategy $\boldsymbol{\pi}_{new}$ for the next segment. This feature-to-strategy loop allows the model to respond to its own generated trends, mimicking human cognitive adjustment during multi-step problem solving without requiring gradient updates during inference.
The number of strategy updates scales as $O(H/K)$, so $K$ directly controls the accuracy-latency trade-off: smaller $K$ adapts more frequently, while larger $K$ reduces forward passes. Because each update consists of deterministic feature extraction, a lightweight MLP, and low-dimensional recursion, the overhead is substantially lighter than gradient-based test-time adaptation. However, very small $K$ can increase latency, and very large $K$ can make the strategy stale under non-stationary dynamics.

\subsubsection{Zero-shot Variable Length Support}
A key advantage of the Feature-to-Strategy paradigm is that its feature extractor and decoder are not tied to a fixed output vector length. The global features ($\boldsymbol{\beta}, \boldsymbol{\alpha}$) are obtained via projection onto basis functions (Chebyshev and Fourier), allowing the same feature definitions to be computed for different context lengths after normalization. Similarly, since FSA generates a strategy rather than a fixed-length vector, it can produce different horizons by extending the autoregressive recursion or the re-estimation loop. This flexibility is not unique to FSA--autoregressive Transformer forecasters can also roll out variable horizons--but in our experimental setup FSA is trained once and reused across horizons, whereas the direct Seq2Seq baselines are trained separately for each target horizon.

\section{Experiments}

In this section, we conduct experiments to evaluate the effectiveness of our proposed framework. The primary goal is to assess the model's zero-shot generalization capability across diverse domains and varying temporal dependencies. We compare our approach against representative task-specific models and time series foundation model architectures under a controlled training setup.

\subsection{Experimental Setup} 
\label{sec:Experimental Setup}
\textbf{Datasets.} To construct a comprehensive pre-training corpus, we curate a large-scale dataset collection spanning multiple domains, including cloud computing logs and traffic flows. Specifically, we utilize Azure VM Traces 2017, Borg Cluster Data 2011, and Alibaba Cluster Trace 2018 from the CloudOpsTSF benchmark\cite{woo2023pushing}. Additionally, we incorporate PEMS07 and Q-Traffic from LibCity\cite{wang2023towards}, as well as the Traffic Hourly dataset from the Monash archive\cite{godahewa2021monash}. We selected 104807 samples from these 6 datasets.

To rigorously evaluate zero-shot performance, we adhere to a strictly disjoint evaluation protocol where test datasets are unseen during training. We benchmark on seven widely-used real-world datasets: ETT(ETTh1, ETTh2, ETTm1, ETTm2)~\cite{yu2018long}, Electricity~\cite{ECL}, Exchange Rate~\cite{lai2018modeling}, and Weather~\cite{weather}. More details about the datasets we used are listed in Appendix \ref{datasets_details}.

\textbf{Baselines.} We compare our method with two categories of baselines:
\begin{itemize}
    \item \textbf{Task-Specific Models:} We include the canonical Transformer and the patch-based model PatchTST~\cite{nie2022time} to establish performance baselines under standard supervised settings, this model is also widely used to build foundation models like panda~\cite{lai2025panda}.
    \item \textbf{Time Series Foundation Models:} To demonstrate the advantage of our autoregressive architecture, we compare with the architecture of the leading foundation models, including TimesFM~\cite{das2024decoder}, Chronos~\cite{ansari2024chronos}.   
\end{itemize}
All neural baselines are trained from scratch on the same source corpus as FSA. This is a controlled architecture comparison under matched data and comparable model size; it is not intended to claim superiority over full-scale industrial TSFMs trained on much larger proprietary or public corpora.

\begin{table*}[t]
\centering
\caption{Full zero-shot multivariate forecasting results (measured by MSE) across all datasets. The input sequence length is fixed to 96, while the output horizons include 24, 48, and 96. Sequence-to-sequence models are trained separately for each target forecasting horizon. In contrast, FSA is trained only once using an autoregressive step size of 3.}
\label{tab:zeroshot_avg_mse}
\begin{small}
\begin{tabular}{lcccccccccccc}
\toprule
\multirow{3}{*} {Datasets} 
& \multicolumn{2}{c}{FSA} 
& \multicolumn{2}{c}{TimesFM} 
& \multicolumn{2}{c}{TimesFM2.5}
& \multicolumn{2}{c}{PatchTST} 
& \multicolumn{2}{c}{Chronos}
& \multicolumn{2}{c}{Transformer} \\
\cmidrule(lr){2-3} \cmidrule(lr){4-5} \cmidrule(lr){6-7} 
\cmidrule(lr){8-9} \cmidrule(lr){10-11} \cmidrule(lr){12-13}
& MSE & MAE & MSE & MAE & MSE & MAE & MSE & MAE & MSE & MAE & MSE & MAE \\
\midrule

\multicolumn{13}{c}{\textbf{Horizon = 24}} \\
\midrule
ETTh1        & 0.456 & 0.439 & 0.702 & 0.536 & 0.639 & 0.509 & 0.487 & 0.448 & 0.492 & 0.463 & 1.069 & 0.676 \\
ETTh2        & 0.198 & 0.286 & 0.231 & 0.311 & 0.193 & 0.286 & 0.222 & 0.300 & 0.232 & 0.301 & 0.407 & 0.428 \\
ETTm1        & 0.562 & 0.426 & 0.447 & 0.393 & 0.569 & 0.441 & 0.606 & 0.470 & 0.790 & 0.493 & 1.608 & 0.817 \\
ETTm2        & 0.121 & 0.219 & 0.122 & 0.215 & 0.120 & 0.220 & 0.133 & 0.227 & 0.137 & 0.226 & 0.309 & 0.388 \\
Electricity  & 0.318 & 0.416 & 0.757 & 0.699 & 0.686 & 0.630 & 0.397 & 0.478 & 0.412 & 0.491 & 1.205 & 0.854 \\
Exchange     & 0.033 & 0.126 & 0.030 & 0.117 & 0.029 & 0.118 & 0.034 & 0.128 & 0.034 & 0.124 & 0.142 & 0.279 \\
Weather      & 0.120 & 0.159 & 0.119 & 0.145 & 0.123 & 0.156 & 0.121 & 0.160 & 0.123 & 0.155 & 0.247 & 0.264 \\
\midrule
Average      & \textbf{0.258} & \textbf{0.296} & 0.344 & 0.345 & 0.337 & 0.337 & 0.286 & 0.316 & 0.317 & 0.322 & 0.712 & 0.529 \\

\midrule
\multicolumn{13}{c}{\textbf{Horizon = 48}} \\
\midrule
ETTh1        & 0.495 & 0.463 & 0.711 & 0.546 & 0.661 & 0.522 & 0.510 & 0.467 & 0.614 & 0.514 & 0.550 & 0.449 \\
ETTh2        & 0.256 & 0.322 & 0.347 & 0.376 & 0.251 & 0.322 & 0.319 & 0.351 & 0.346 & 0.364 & 0.410 & 0.404 \\
ETTm1        & 0.731 & 0.490 & 1.073 & 0.621 & 0.750 & 0.527 & 1.257 & 0.638 & 1.091 & 0.571 & 1.848 & 0.863 \\
ETTm2        & 0.166 & 0.256 & 0.226 & 0.301 & 0.164 & 0.263 & 0.196 & 0.277 & 0.207 & 0.278 & 0.341 & 0.400 \\
Electricity  & 0.362 & 0.445 & 0.798 & 0.701 & 0.703 & 0.638 & 0.457 & 0.520 & 0.633 & 0.632 & 0.373 & 0.384 \\
Exchange     & 0.068 & 0.180 & 0.061 & 0.170 & 0.052 & 0.160 & 0.062 & 0.176 & 0.071 & 0.180 & 0.090 & 0.221 \\
Weather      & 0.177 & 0.221 & 0.191 & 0.213 & 0.168 & 0.206 & 0.184 & 0.223 & 0.193 & 0.230 & 0.295 & 0.294 \\
\midrule
Average      & \textbf{0.322} & \textbf{0.340} & 0.487 & 0.418 & 0.393 & 0.377 & 0.426 & 0.379 & 0.451 & 0.396 & 0.558 & 0.431 \\

\midrule
\multicolumn{13}{c}{\textbf{Horizon = 96}} \\
\midrule
ETTh1        & 0.543 & 0.492 & 1.067 & 0.692 & 0.695 & 0.548 & 0.573 & 0.500 & 0.807 & 0.622 & 0.661 & 0.524 \\
ETTh2        & 0.312 & 0.355 & 0.574 & 0.500 & 0.330 & 0.372 & 0.451 & 0.417 & 0.675 & 0.484 & 0.542 & 0.460 \\
ETTm1        & 0.770 & 0.511 & 1.252 & 0.701 & 0.860 & 0.581 & 1.882 & 0.774 & 2.108 & 0.799 & 1.662 & 0.805 \\
ETTm2        & 0.215 & 0.292 & 0.390 & 0.405 & 0.217 & 0.302 & 0.279 & 0.327 & 0.432 & 0.379 & 0.362 & 0.405 \\
Electricity  & 0.403 & 0.477 & 1.261 & 0.876 & 0.747 & 0.667 & 0.525 & 0.562 & 0.878 & 0.765 & 0.531 & 0.500 \\
Exchange     & 0.127 & 0.246 & 0.248 & 0.364 & 0.102 & 0.223 & 0.125 & 0.245 & 0.174 & 0.276 & 0.226 & 0.349 \\
Weather      & 0.231 & 0.270 & 0.353 & 0.341 & 0.223 & 0.260 & 0.253 & 0.280 & 0.323 & 0.310 & 0.305 & 0.318 \\
\midrule
Average      & \textbf{0.372} & \textbf{0.378} & 0.735 & 0.554 & 0.453 & 0.422 & 0.584 & 0.444 & 0.771 & 0.519 & 0.613 & 0.480 \\

\bottomrule
\end{tabular}
\end{small}
\vskip -0.2in
\end{table*}

\textbf{Implementation Details.} All models are trained from scratch on the same datasets and evaluated in a zero-shot manner on disjoint test datasets. Baseline models retain their original architectural design, but widths, depths, and heads are reduced to comparable capacity. We use context length $T=96$ during pretraining and evaluate horizons $H\in\{24,48,96\}$. FSA uses AR order $p=3$ and re-estimation interval $K=3$ unless otherwise stated. All runs use batch size 64, AdamW with learning rate $10^{-4}$, weight decay 0.01, cosine decay with 10\% warmup, gradient clipping at 1.0, and a 90/10 train-validation split. Early stopping uses validation MSE with patience 10 and minimum delta $10^{-4}$. More implementation details can be found in Appendix \ref{sec:Optimization and Implementation Details}.

\textbf{Zero-Shot Forecasting}. FSA is designed for univariate forecasting. For multivariate benchmark datasets, we follow a dimension-wise decomposition protocol: each variable dimension is treated as an independent univariate forecasting instance and predictions are performed separately. This protocol evaluates marginal per-series forecasting quality under a common benchmark setting, but it does not test cross-series dependency modeling. We therefore do not claim that FSA exploits multivariate interactions.

\subsection{Results}

Table \ref{tab:zeroshot_avg_mse} summarizes the zero-shot forecasting performance across diverse prediction horizons. Under our controlled pretraining and capacity-matched setting, FSA obtains the best average MSE across all evaluated horizons ($H \in \{24, 48, 96\}$) without task-specific fine-tuning. This supports the effectiveness of feature-to-strategy learning in the tested low-to-moderate capacity regime.

\textit{Short-term Precision and Competitive Margins}. At the $H=24$ horizon, FSA achieves a Mean Squared Error (MSE) of 0.258, representing a significant improvement over established foundation models such as TimesFM (0.344) and Chronos (0.317). Notably, FSA maintains a distinct advantage over PatchTST (0.286), suggesting that our distribution-aware synthesis provides a more precise prior for short-term dynamics than traditional patching mechanisms.

\textit{Robustness to Long-Horizon Error Accumulation}. The performance margin between FSA and the baselines widens substantially as the forecasting horizon extends. At $H=48$, FSA reduces the average MSE to 0.322, whereas the best-performing baseline (TimesFM 2.5) lags at 0.393. At $H=96$, while sequence-to-sequence models like TimesFM and Chronos suffer from severe performance decay—exceeding an MSE of 0.7—FSA maintains high predictive fidelity with an MSE of 0.372.

This widening performance gap suggests that FSA is less sensitive to error accumulation in this setting. We attribute this behavior to the use of compact structural descriptors and periodically refreshed autoregressive strategies, while noting that longer-context and longer-horizon benchmarks remain important future tests.

\begin{table}[t]
\centering
\caption{Inference cost of autoregressive re-estimation on seven datasets ($T=96$, $H=24$, batch size 32). Times are averages over three timed runs after one warmup.}
\label{tab:reestimate_cost}
\begin{small}
\begin{tabular}{lcc}
\toprule
Re-estimation interval $K$ & Time (s) & Avg MSE \\
\midrule
1 & 142.60 & 0.178 \\
3 & 70.13 & \textbf{0.177} \\
5 & 62.05 & 0.181 \\
8 & 50.10 & 0.192 \\
10 & 48.61 & 0.193 \\
\midrule
Transformer & 56.06 & 0.263 \\
\bottomrule
\end{tabular}
\end{small}
\end{table}

Table~\ref{tab:reestimate_cost} reports the cost of the re-estimation loop. The default $K=3$ provides the best accuracy in this setting, while larger intervals reduce latency at the cost of less frequent adaptation.

\subsection{Zero-Shot Generalization under Data-Constrained Regimes}

While Section \ref{sec:Experimental Setup} evaluates performance on large-scale, diverse datasets, this section investigates model robustness under constrained training regimes. Here, we employ a sequential sliding-window strategy where the model is exposed only to a localized segment of the underlying time series distribution. This setup probes the model's ability to extract transferable structural invariants from limited and low-diversity training signals—a critical requirement for zero-shot generalization in data-scarce environments.
\begin{table}[ht]
\caption{Univariate zero-shot forecasting performance (MSE). Models are evaluated on a forecasting horizon of $H=24$ with an input length of $L=96$ across varying training sample sizes (10K, 20K, and 50K).}
\label{tab:mse_small_scale}
\resizebox{\linewidth}{!}{
\begin{tabular}{lccc|ccc}
\toprule
\textbf{Dataset} 
& \multicolumn{3}{c}{\textbf{FSA}} 
& \multicolumn{3}{c}{\textbf{PatchTST}} \\
\cmidrule(lr){2-4}\cmidrule(lr){5-7}
 & 10K & 20K & 50K & 10K & 20K & 50K \\
\midrule
ETTh1          & 0.039 & 0.038 & 0.033 & 0.043 & 0.044 & 0.041 \\
ETTh2          & 0.154 & 0.131 & 0.141 & 0.250 & 0.194 & 0.177 \\
ETTm1          & 0.018 & 0.019 & 0.014 & 0.017 & 0.017 & 0.018 \\
ETTm2          & 0.108 & 0.114 & 0.110 & 0.181 & 0.155 & 0.198 \\
electricity    & 0.872 & 0.893 & 0.846 & 2.019 & 1.518 & 1.337 \\
exchange\_rate & 0.046 & 0.048 & 0.030 & 0.039 & 0.040 & 0.039 \\
weather        & 0.0007& 0.0006& 0.0005& 0.0006& 0.0006& 0.0006\\
\midrule
\textbf{Avg}   & \textbf{0.177} & \textbf{0.177} & \textbf{0.168}
               & \textbf{0.364} & \textbf{0.281} & \textbf{0.259} \\
\bottomrule
\end{tabular}
}
\end{table}

\paragraph{Results \& Analysis.} 
Table \ref{tab:mse_small_scale} reports the univariate zero-shot forecasting results under this restricted training regime. Compared to the cross-dataset pre-training setting in Section 5.1, this scenario induces significant distributional scarcity, presenting a more rigorous test of model architecture.

\begin{itemize}
    \item \textbf{Robustness to Restricted Distributions}: FSA consistently outperforms PatchTST across nearly all datasets and scales. The performance gap is most pronounced in datasets characterized by high non-stationarity and complex dynamics, such as Electricity and ETTh2. For instance, at the 10K scale on the Electricity dataset, FSA reduces MSE by over 56\% compared to PatchTST, suggesting that FSA is better equipped to capture latent temporal structures when training diversity is localized.
    \item \textbf{Superior Sample Efficiency}: Notably, FSA achieves an average MSE of 0.177 with only 10K samples—surpassing PatchTST's performance even when the latter is provided with 50K samples (0.259). This five-fold improvement in sample efficiency indicates that FSA's inductive bias is significantly more effective at regularizing the model against overfitting to local training segments.
    \item \textbf{Scaling Stability}: As the training volume scales from 10K to 50K, FSA exhibits highly stable performance trajectories. While PatchTST shows a steeper improvement curve, it remains unable to bridge the performance gap, highlighting FSA's inherent resilience to the noise and limited coverage typical of small-scale sequential training.
\end{itemize}

\subsection{Ablation Study}

To isolate the individual contributions of the architectural components within FSA, we perform a systematic ablation study by selectively removing the global and local feature modules. To ensure a rigorous comparison, all variants are pre-trained under identical configurations and evaluated using the standard zero-shot forecasting protocol.

\begin{table}[ht]
\centering
\caption{Ablation analysis of feature components. Reported values are averages across all benchmark datasets at $H=24$.}
\label{tab:ablation_avg}
\begin{small}
\begin{tabular}{lcc}
\toprule
Model Variant & MSE & MAE \\
\midrule
Full (Global + Local) & \textbf{0.258} & \textbf{0.296} \\
w/o Local Features & 0.286 & 0.311 \\
w/o Global Features & 0.407 & 0.375 \\
\bottomrule
\end{tabular}
\end{small}
\end{table}

\begin{table}[ht]
\centering
\caption{Ablation of the two main design choices in the 20K-data setting ($T=96$, $H=24$).}
\label{tab:raw_direct_ablation}
\begin{small}
\begin{tabular}{lc}
\toprule
Setting & Avg MSE \\
\midrule
FSA (full) & \textbf{0.177} \\
Replace strategy head with direct prediction & 0.206 \\
Remove feature engineering (raw input) & 0.387 \\
\bottomrule
\end{tabular}
\end{small}
\end{table}

\begin{figure}[ht]
    \centering
    \includegraphics[width=1\linewidth]{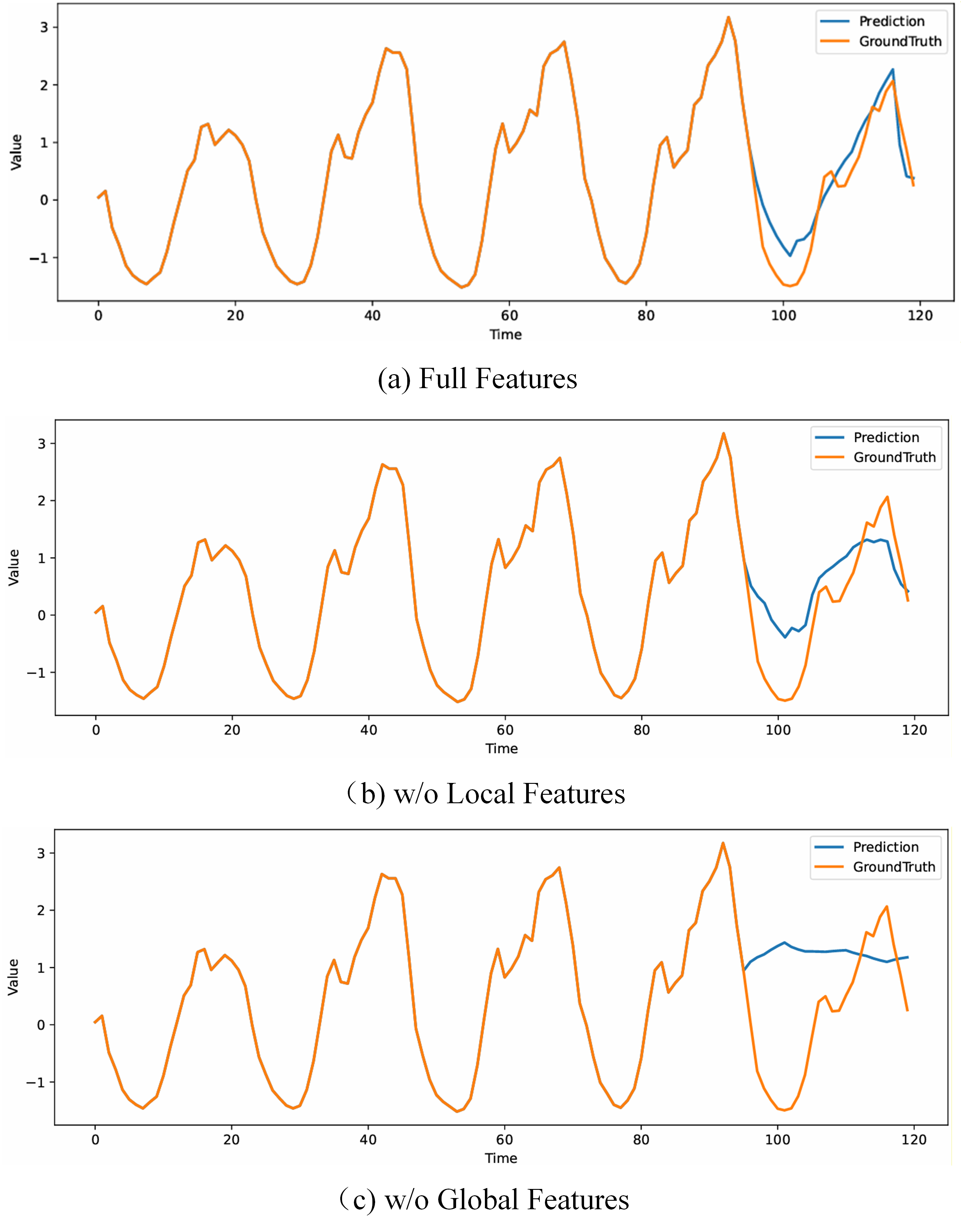}
    \caption{Visualization of ablation study results. (a) By integrating local and global features, the model effectively captures both local dynamics and global periodicity. (b) With global features only, the model captures coarse-grained periodic patterns but misses local variations. (c) Relying solely on local features, the model fails to capture trends and tends to produce mean-value predictions.}
    \label{fig:ablation}
\end{figure}

\paragraph{Results \& Analysis.} 
As summarized in Table \ref{tab:ablation_avg}, the omission of either component induces a clear performance degradation, confirming their complementary roles. Notably, the removal of global features results in a severe performance penalty, with MSE increasing by 57.8\%. This suggests that global descriptors serve as the primary structural prior; without them, the model loses the ability to anchor its predictions within the broader temporal context, leading to a breakdown in zero-shot generalization.

The model's predictions collapse toward smoothed or mean-value trajectories, as shown in Figure \ref{fig:ablation}(c), and fail to capture both trend and periodic structure. This behavior suggests that local descriptors, while informative for short-term variations, lack the structural context necessary for stable autoregressive strategy generation in a zero-shot setting. The full model, which integrates both global and local features, consistently achieves the lowest error across all datasets. As shown in Figure \ref{fig:ablation}(a), this integration enables FSA to simultaneously capture global periodicity and local temporal variations, leading to accurate and stable long-horizon predictions. These results demonstrate that global features provide essential structural priors, while local features refine the strategy to adapt to immediate dynamics. Their combination is therefore critical for learning transferable and interpretable forecasting policies.

Table~\ref{tab:raw_direct_ablation} further separates the effect of feature engineering from the effect of strategy-based decoding. Replacing the strategy head with direct prediction increases average MSE from 0.177 to 0.206, while removing structured features and using raw input increases MSE to 0.387. This indicates that FSA's gain is not due to the reformulation alone; it comes from the combination of structured temporal descriptors and a compact autoregressive strategy space. Table~\ref{tab:reestimate_cost} shows that moderate order $p=3$ performs best, while larger orders add latency and can overfit local fluctuations.

\subsection{Strategy Visualization}

One key advantage of FSA lies in its explicit and interpretable forecasting strategy. Unlike sequence-to-sequence models that implicitly encode forecasting behavior within high-dimensional hidden states, FSA predicts a low-dimensional autoregressive strategy whose parameters directly govern the forecasting dynamics. To further understand how the model adapts its behavior across different datasets and horizons, we visualize and analyze the learned autoregressive coefficients.

\begin{figure}
    \centering
    \includegraphics[width=1\linewidth]{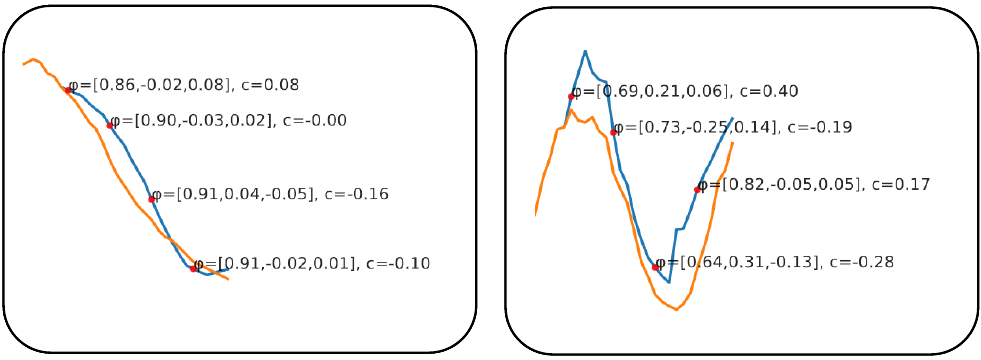}
    \caption{Visualization of AR and const parameter: Autoregressive dynamics at different sequences }
    \label{fig:strategy}
    \vskip -0.1in
\end{figure}

In Figure~\ref{fig:strategy}, we visualize representative forecasting segments together with the corresponding autoregressive coefficients predicted by FSA at each forecasting step. Each marker denotes the re-estimated AR parameters and constant used for subsequent prediction, illustrating how the strategy evolves as new observations are incorporated. We observe that the learned AR coefficients exhibit clear structure and meaningful variation across different temporal patterns. In smoother and more monotonic regimes (left panel), the model assigns relatively larger weights to the most recent lag, producing a stable and gradually decaying autoregressive profile. This behavior encourages smooth extrapolation and prevents overreaction to local noise. In contrast, in regions with sharper turning points or rapid fluctuations (right panel), the predicted coefficients redistribute mass across multiple lags, allowing the model to respond more flexibly to short-term dynamics. 

\section{Limitations}
\label{sec:limitations}

FSA is designed as a zero-shot univariate forecasting framework, and its current form does not model cross-series dependencies. On multivariate benchmarks, our evaluation decomposes each channel into an independent univariate instance; therefore, the results measure marginal forecasting quality rather than native multivariate reasoning. Extending the feature-to-strategy idea to strategies that share information across variables is an important future direction.

The current decoder is also intentionally compact. Although nonlinear structure enters through feature extraction and the MLP strategy generator, the final rollout is an AR-style linear recursion. This makes the forecasting policy interpretable and efficient, but it can underfit highly irregular high-frequency signals, chaotic dynamics, abrupt regime shifts, very long-memory processes, or systems whose future evolution depends on exogenous events not visible in the input window. The re-estimation loop provides a practical adaptation mechanism for moderate non-stationarity, but it is not a guarantee under structural breaks.

\section{Conclusion}

We present FSA, a novel feature-to-strategy framework that shifts the zero-shot time-series forecasting paradigm from direct observation prediction to the synthesis of autoregressive strategies. By disentangling global structural invariants from local temporal dynamics, FSA incorporates robust inductive biases that facilitate transferable temporal reasoning across diverse distributions.

Our evaluations show that FSA outperforms capacity-matched Transformer-based architectures under controlled pretraining and constrained-data regimes, achieving strong generalization and sample efficiency in the tested setting. These results suggest that zero-shot forecasting can benefit not only from scaling model size and data, but also from revisiting the learning objective. By prioritizing strategy generation over point-wise value prediction, FSA offers an interpretable and data-efficient direction for future forecasting architectures.

\section*{Impact Statement}

This paper presents work whose goal is to advance the field of Time Series Foundation Model. There are many potential societal consequences of our work, none which we feel must be specifically highlighted here.

\bibliography{example_paper}
\bibliographystyle{icml2026}

\newpage
\appendix
\onecolumn
\section{Training Details}

\subsection{Optimization and Implementation Details}
\label{sec:Optimization and Implementation Details}
For all baselines, we adopt the same data pipeline as our method. Each sample consists of a context window of length $T=96$ and a prediction horizon $H\in\{24,48,96\}$. Direct sequence-to-sequence baselines are trained separately for each target horizon, while FSA is trained once with AR order $p=3$ and uses iterative rollout to produce all reported horizons. We optimize all models using AdamW~\cite{loshchilov2018decoupled} with learning rate $10^{-4}$, weight decay 0.01, a cosine learning-rate schedule, 10\% linear warmup, batch size 64, and gradient clipping at 1.0. Experiments use a 90/10 train-validation split of the pretraining corpus; validation MSE is used for early stopping and model selection.

\textbf{Model Configurations:} FSA uses 22 input features, a 3-layer MLP strategy generator with hidden size 512, GELU activations, dropout 0.1, and LayerNorm after each linear layer. The output dimension is $p+q+2=6$ for the default ARIMA-inspired $(p,d,q)=(3,0,1)$ parameterization. The capacity-matched baselines use the following configurations: PatchTST ($d=128$, heads=16, layers=3), TimesFM 2.5 ($d=512$, heads=8, layers=6), TimesFM 1.0 ($d=128$, heads=8, layers=3), Transformer ($d=128$, heads=16, encoder layers=2, decoder layers=1), and Chronos 2 ($d=128$, layers=3, heads=8).

\textbf{Learning Rate Scheduling:} To facilitate stable convergence in the non-linear strategy space, we employ a \textbf{CosineAnnealingLR} scheduler. The learning rate $\eta$ decays from its initial value to a minimum $\eta_{\min}$, allowing for aggressive exploration in early stages and fine-grained refinement in later epochs.

\textbf{Stability Mechanisms:} To mitigate the risk of vanishing or exploding gradients inherent in recursive unrolling, we implement two safeguards:
\begin{itemize}
    \item \textbf{Gradient Clipping}: All gradients are clipped to a maximum norm of $1.0$.
    \item \textbf{Strategy Constraints}: As described in Section~3.2, the $\tanh$ activation bounds the AR coefficients $\phi_i$ within a stable empirical range, reducing the risk that recursive unrolling diverges. This constraint improves rollout stability in practice, but it is not a full stationarity guarantee for every AR($p$) process.
\end{itemize}

\textbf{Regularization:} Early stopping is conducted with a patience of 10 epochs and minimum delta $10^{-4}$ based on validation MSE. The best-performing model checkpoint is selected for final evaluation on unseen zero-shot benchmarks.

\section{Additional Experimental Details}
\subsection{Pretraining Datasets}
\label{datasets_details}
Rather than relying on extremely large and heterogeneous corpora, our goal is to evaluate whether a model trained on a \emph{limited and controlled set of source domains} can generalize to unseen domains in a zero-shot manner. To this end, we intentionally construct a compact pretraining corpus covering a small number of representative domains, and evaluate generalization on entirely disjoint datasets from different application areas.

Specifically, we collect time series data from six datasets spanning two source domains: cloud computing workloads and traffic systems. From the CloudOpsTSF benchmark~\cite{woo2023pushing}, we include Azure VM Traces 2017, Borg Cluster Data 2011, and Alibaba Cluster Trace 2018, which capture resource utilization dynamics in large-scale distributed systems. In addition, we incorporate traffic-related datasets from LibCity~\cite{wang2023towards}, including PEMS07 and Q-Traffic, as well as the Traffic Hourly dataset from the Monash Time Series Forecasting Archive~\cite{godahewa2021monash}. We use these datasets via gifteval~\cite{aksu2024gift} hugging face repository. To avoid redundancy and enhance training diversity, we sample only one instance from each series. 

We conduct Zero-shot experiments on four benchmark time series datasets: \textbf{ETT}~\cite{yu2018long}, \textbf{Electricity}~\cite{ECL}, \textbf{Exchange Rate}~\cite{lai2018modeling}, and \textbf{Weather}~\cite{weather}. The ETT dataset contains data collected from electricity transformers, including load and oil temperature measurements recorded every 15 minutes from July 2016 to July 2018. The Electricity dataset consists of the hourly electricity consumption of 321 customers between 2012 and 2014. The Exchange\_rate dataset records the daily exchange rates of eight foreign countries spanning from 1990 to 2016. The Weather dataset includes meteorological observations recorded every 10 minutes throughout the year 2020, containing 21 weather-related indicators such as air temperature, humidity, and wind speed. 

\subsection{Baseline models}
We use baseline models from these repositories: 

https://github.com/google-research/timesfm

https://github.com/amazon-science/chronos-forecasting

https://github.com/thuml/Time-Series-Library/

\begin{table}[h]
\centering
\caption{Summary of datasets used for pretraining.}
\label{tab:dataset_summary}
\setlength{\tabcolsep}{6pt}
\begin{small}
\begin{tabular}{cccc}
\toprule
Dataset & Domain & Time Granularity & Samples\\
\midrule
Azure VM Traces 2017 & Cloud Computing & Variable & 33333\\
Borg Cluster Data 2011 & Cloud Computing & Variable & 3064\\
Alibaba Cluster Trace 2018 & Cloud Computing & Variable & 33333\\
PEMS07 & Traffic & 5 minutes & 882\\
Q-Traffic & Traffic & 15 minutes & 33333\\
Traffic Hourly (Monash) & Traffic & 1 hour & 862\\
\midrule
Total & -- & -- & 104807 \\
\bottomrule
\end{tabular}
\end{small}
\end{table}

\begin{figure}[ht]
    \centering
    \includegraphics[width=1\linewidth]{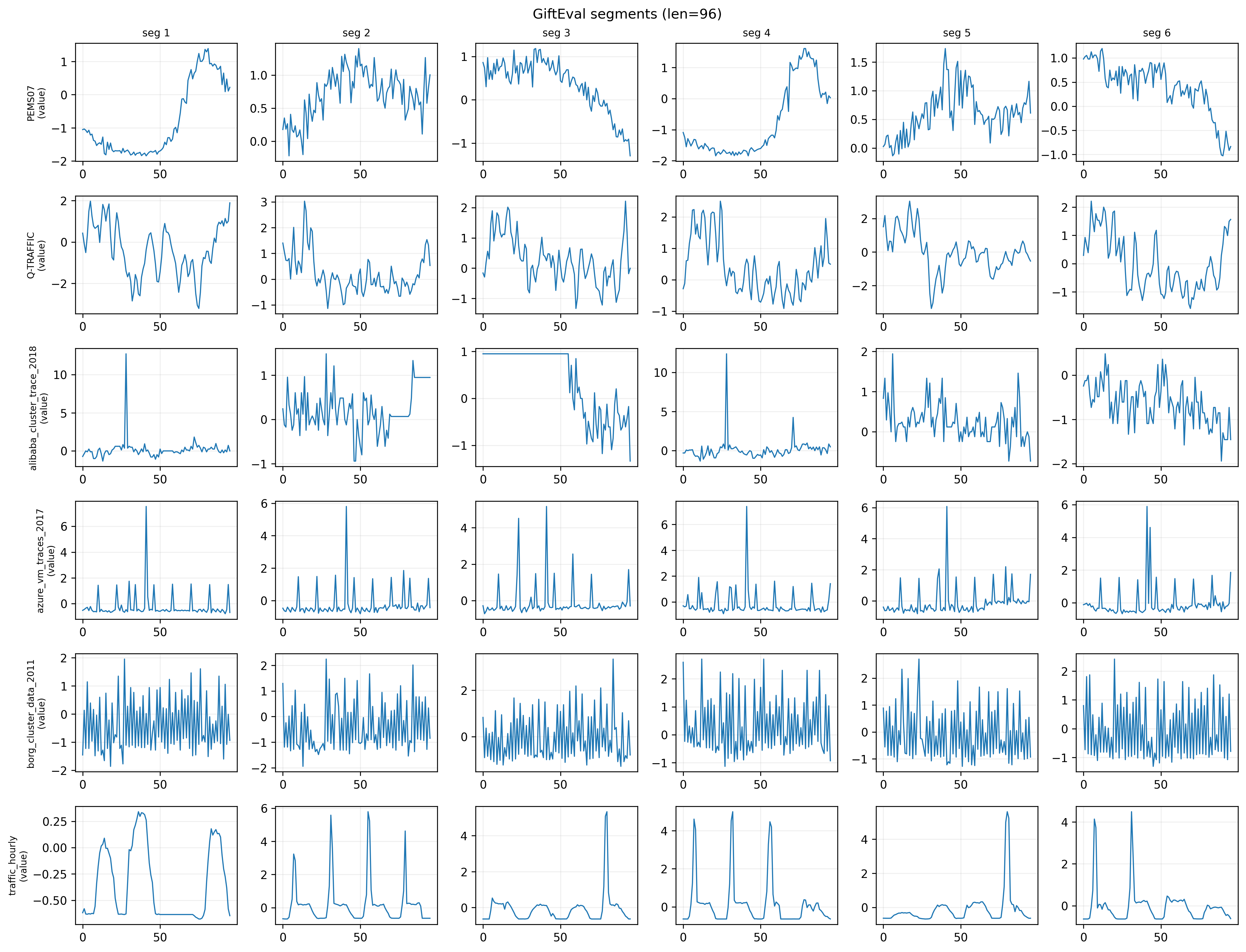}
    \caption{Visualization of pretrain datasets}
    \label{fig:pretrain_gallery}
\end{figure}

\section{Ablation results}
\subsection{full results}

\begin{table}[ht]
\centering
\caption{Ablation study on global and local feature components.}
\label{tab:ablation_full}
\begin{small}
\begin{tabular}{c|cc|cc|cc}
\toprule
\multirow{2}{*}{Dataset}& \multicolumn{2}{c|}{Full (Global + Local)}
& \multicolumn{2}{c|}{w/o Local}
& \multicolumn{2}{c}{w/o Global} \\
& MSE & MAE & MSE & MAE & MSE & MAE \\
\midrule
ETTh1 & \textbf{0.456} & \textbf{0.439} & 0.507 & 0.454 & 0.781 & 0.548 \\
ETTh2 & \textbf{0.198} & \textbf{0.286} & 0.211 & 0.297 & 0.228 & 0.318 \\
ETTm1 & \textbf{0.562} & \textbf{0.426} & 0.582 & 0.434 & 0.559 & 0.441 \\
ETTm2 & \textbf{0.121} & \textbf{0.219} & 0.126 & 0.222 & 0.130 & 0.233 \\
Electricity & \textbf{0.318} & \textbf{0.416} & 0.422 & 0.497 & 0.979 & 0.779 \\
Exchange Rate & 0.033 & 0.126 & \textbf{0.031} & \textbf{0.122} & 0.036 & 0.127 \\
Weather & \textbf{0.120} & 0.159 & 0.121 & \textbf{0.153} & 0.135 & 0.176 \\
\midrule
Average & \textbf{0.258} & \textbf{0.296} & 0.286 & 0.311 & 0.407 & 0.375 \\
\bottomrule
\end{tabular}
\end{small}
\end{table}

\end{document}